# 3D Reconstruction of deciduous Trees using low-cost UAV- and Crane-based Photogrammetry for Monitoring Shoot Elongation across entire Canopies

Stephan Nebiker[1], Micha Tschanz[1], Nando Amport[1], Frederik Baumgarten[2]

[1] Institute of Geomatics, FHNW University of Applied Sciences and Arts Northwestern Switzerland, 4132 Muttenz, Switzerland – stephan.nebiker@fhnw.ch; micha.tschanz@fhnw.ch; nando.amport@gmail.com

[2] Physiological Plant Ecology group, University of Basel, 4056 Basel – frederik.baumgarten@unibas.ch

**Abstract**

Tree growth determines how much CO2 is sequestered from the atmosphere and temporarily stored in woody biomass. At the same time tree growth is affected by increasing temperatures, more frequent drought periods, late frosts and other extreme events associated with climate change. While continuous measurements of radial (secondary) tree growth using dendrometers are well established, monitoring of shoot elongation (primary growth) has largely been neglected because suitable measurement techniques are lacking. As a result, the effects of climate change on primary tree growth remain insufficiently understood. This work aims at reconstructing native deciduous trees in 3D as a basis for measuring and monitoring shoot elongation over entire tree canopies. Here we explored the use of low-cost UAV photogrammetry and of a multi-camera CraneCam system under real-world conditions. Data were collected in two study areas over an entire growing season. We present sensor evaluations, photogrammetric data acquisition and processing strategies. A special focus is placed on the analysis of the resulting photogrammetric 3D point clouds in terms of accuracy, resolution and completeness. Results demonstrate 3D point accuracies of 5-6 mm for entire trees using consumer-grade UAVs weighing less than 250 g and a 3D reconstruction completeness between 92% and 98% depending on the UAV type. The paper introduces a novel 3D-printed ground-truth branch to evaluate the capability to reconstructing fine-detail structures such as thin tree shoots. Finally, we discuss operational challenges and initial experiments towards a skeletonization of entire trees based on photogrammetric point clouds.

## 1. Introduction

Every spring, deciduous forests turn green again within a few days to weeks. This rapid leaf emergence is enabled by the preformation and overwintering of leaf tissue in frost-resistant buds. In many species, such as maple and ash, all leaf structures that develop during the growing season are already present in these buds. This strategy is known as 'determined growth'. Other species, such as birch and poplar, can continue forming new tissue during the same growing season ('indeterminate growth'). When and to what extent trees grow determines how much CO2 is removed from the atmosphere and stored in the form of wood. Tree growth is also affected by climate change: Increased temperatures, more frequent drought periods, late spring frosts and other extreme events are putting trees under increasing stress (Strauss & Fischer, 2025). Tree species with flexible growth behaviour may be better to cope with future conditions, as they can adjust their growing season more readily to variable environments. Trees with extended growth periods or multi-phase shoots are therefore of particular interest in studies of adaptability (Braun et al., 2023; Strauss & Fischer, 2025). However, while radial growth can be monitored continuously using dendrometers, suitable methods for monitoring shoot elongation at whole-tree scale remain limited. Shoot elongation, i.e. longitudinal growth at the tips of current-season shoots, has so far mainly been recorded manually using calipers and rulers (Schiestl-Aalto et al., 2013; Riikonen et al., 2011). Yet, although primary growth typically reaches one to several decimetres per season, accurate integrated measurements at crown scale are still lacking. This paper addresses this measurement problem and aims to contribute towards answering two main questions:

1) When does shoot elongation take place during the growing season (temporal pattern of primary growth)?

2) How much primary growth occurs in absolute terms over parts of, or the entire growing season?

Here we investigate 3D reconstruction of deciduous trees as a basis for subsequent analysis of shoot elongation using crane-based and UAV photogrammetry. The main contributions of our paper include:

- Monitoring over an entire growing season using two different measurement setups: a (low-cost) drone-based setup with terrestrial laser scanning (TLS) data as reference and a setup using crane-based cameras typically used for construction site monitoring.
- A thorough investigation of the characteristics of the UAV cameras used and their effects on point-cloud accuracy, noise and completeness.
- The introduction of a ground-truth branch (GTB), a 3D-printed artificial shoot with known geometry to assess the accuracy and limitations of the detailed reconstruction of tree shoots.
- First results from a whole-tree skeletonization using our UAV-based point clouds.

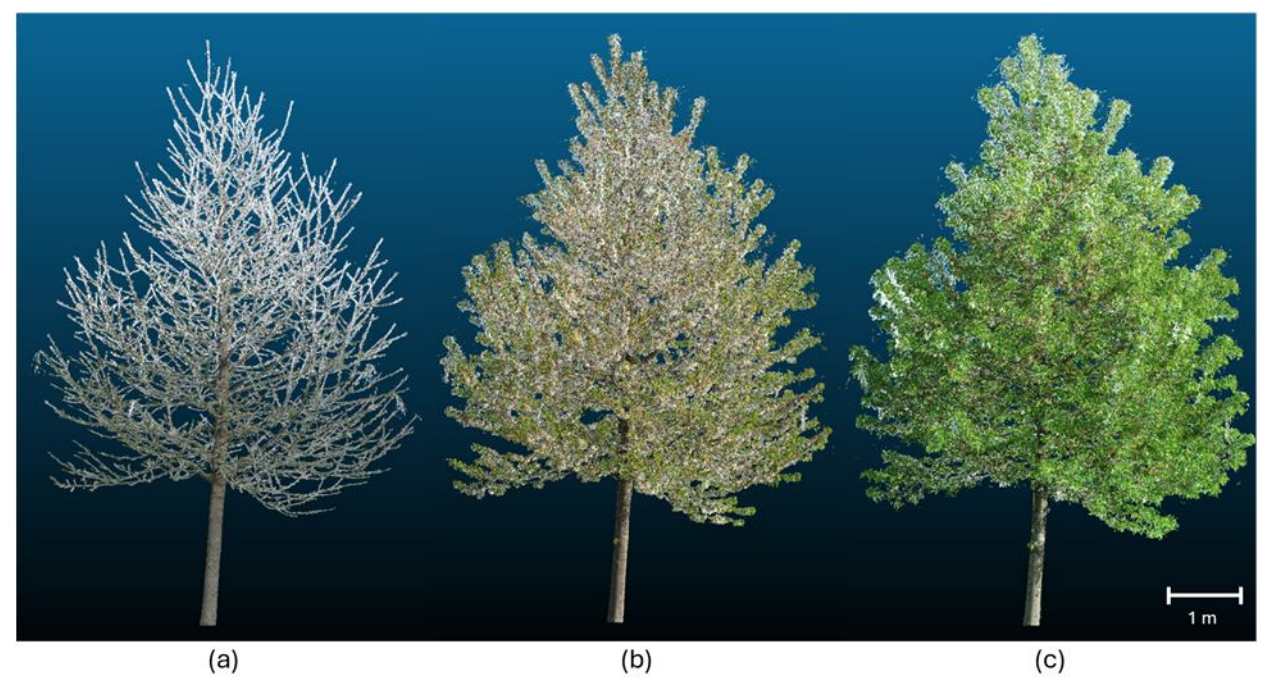

Figure 1. Point clouds of an ornamental cherry tree reconstructed from low-cost UAV imagery taken at three different epochs during the growing season of 2025.

## 2. Related Work

Research for the derivation of various products in the field of forests using terrestrial laser scanning (TLS) is already very advanced and applicable compared to other measurement methods (Beland et al., 2019). Several studies show that TLS can provide high-resolution point clouds of trees with just three to four scanning stations, which are suitable for deriving various tree parameters (Böhme et al., 2024; Bournez et al., 2019; Wang et al., 2024).

UAV-based laser scanning (ULS) approaches, on the other hand, are much less explored. ULS is particularly suitable for single tree detection and crown segmentation (Jaakkola et al., 2017; Xu et al., 2021). However, these methods reach their limits in the case of irregular treetops and multi-stemmed trees (Xu et al., 2021).

Compared to laser scanning methods, photogrammetric approaches for recording forest and tree structures have so far been less researched. Due to the high hardware costs and the large amount of time required for TLS recordings, photogrammetry is a potential alternative. However, photogrammetric point cloud generation is sensitive to lighting conditions and has difficulties with shadows in the image, which limits accuracy (Hartley et al., 2024; Iglhaut et al., 2019).

Scher et al. (2019) focused on the development and validation of a precise method for creating high-resolution 3D models of free-standing individual trees. A DJI Inspire 2 with a Zenmuse X5S camera was used. The trees were flown over with two orbits, each at different heights. The images were not triggered automatically but had to be triggered manually every 2 to 4 seconds. Reference measurements were carried out manually with tape measure and caliper. To measure these references, the trees were felled after recording with the UAVs. According to Scher et al. (2019), high-contrast backgrounds in the photos significantly improve the recognition of terminal shoots. The best shooting conditions are provided by a cloudy sky and a ground covered with snow to be able to see the fine shoots as well as possible.

Many studies have carried out large-scale examinations with SfM photogrammetry. In Zhang et al. (2019), the growth of a single tree, a group of trees and a forest area was calculated using tree crown surfaces. A dense point cloud was generated using the SfM-MVS process to create a crown surface model. The derived crown surfaces of the tree group and the individual tree correlate with the manually measured references with a coefficient of 0.96. The forest section has a correlation coefficient of 0.85. In this work, only the two-dimensional crown surface was analysed for its growth, and no individual shoots were monitored.

Nevertheless, studies such as those by Scher et al. (2019) have also attempted to reconstruct the smallest branch structures. The aim was to use photogrammetry to structurally model individual trees regarding branch structures and branch diameters. Their average root mean square error (RMSE) across all images was 0.8 pixels. For individual GCPs, the maximum pixel error was 1.0 pixels. The resulting point cloud was further used to generate a 3D mesh.

## 3. Test Sites and Reference Data

### 3.1 Test Sites

The park at the campus of FHNW (University of Applied Sciences and Arts Northwestern Switzerland) in Muttenz, Switzerland, served as the main study area. For the investigations, an ornamental cherry (*Prunus serrulata* 'Shirofugen'), a winter lime (*Tilia cordata*) and a Norway maple (*Acer platanoides*) were selected, as these trees have different growth behaviours. The ornamental cherry commonly shows an indeterminate growth, the Norway maple determined growth and the winter lime intermediate growth behaviour. Determined growth is characterized by a single, annually recurring growth flush in spring. In indeterminate growth, leaves and internodes are formed continuously or in intervals throughout the growing season (Lechowicz, 1984). Investigations in the FHNW Park were carried out using UAV photogrammetry and UAV LiDAR.

The second study area is located next to a construction site in the city of Winterthur, Switzerland. In contrast to the park at the FHNW, an entire group of trees with a height of approx. 18 m was examined. The site consisted of typical mixed forest species, with sycamore maple (Acer pseudoplatanus) and European beech (Fagus sylvatica) dominating the stand. The images in Winterthur were recorded using a TEDAMOS CraneCam system from Terradata AG. The system was used to monitor construction site progress (Terradata AG, 2024). Due to the proximity to the group of trees in the study area and a possible use of a CraneCam system on the Swiss Canopy Crane II (Arend & Kahmen, 2019) in the forest laboratory in Hölstein, BL, this additional recording method was also investigated.

### 3.2 Reference Data

To check the quality and completeness of the photogrammetric results, various reference data were recorded and reference objects attached to the investigated trees.

**3.2.1 Terrestrial Laser Scans:** Reference data of the trees in the FHNW Park were recorded using TLS. To collect the reference data, the "Leica RTC360" and the "Trimble X9" scanners were compared. Based on the comparison of the test images, the reference data of the entire measurement series were recorded with the "Leica RTC360", as it offers slightly higher accuracy, lower measurement noise and significantly shorter recording time. The reference measurements were recorded with eight scan stations at a distance of 5.5 m from the tree trunk, so that an accuracy of 2 mm at each point of the tree is guaranteed.

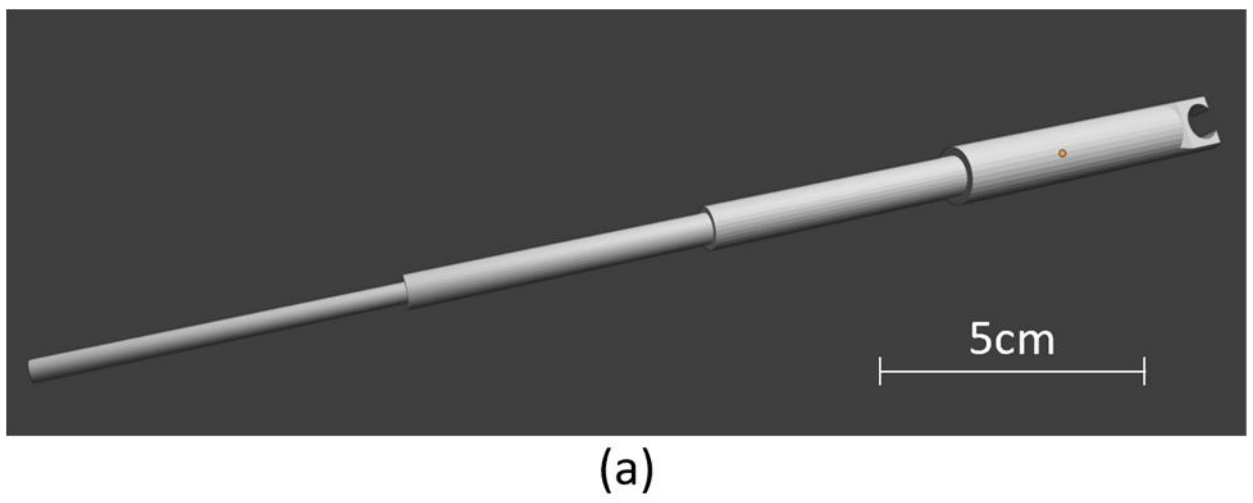

(a)

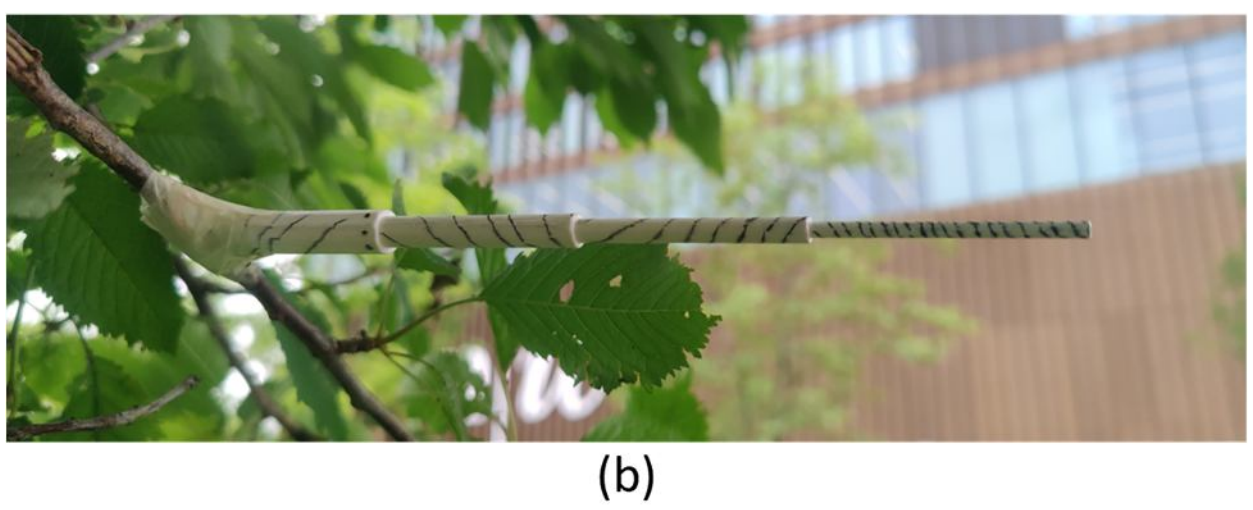

(b)

Figure 2. Modelled ground-truth branch GTB rendered in «Blender» (a) and the 3D-printed version attached to a branch of the ornamental cherry tree (b).

**3.2.2 Ground-Truth Branch (GTB):** To check the TLS reference point cloud and photogrammetrically generated point clouds, a 3D-printed ground-truth branch (GTB) was attached to the tree. 3D printed plant models can be used to check the accuracy of point clouds (Bömer et al., 2024). In this work, an abstracted branch was used as a 3D reference (See Figure 2 (a)). The abstracted branch has four sections with constant diameters (3 mm, 5 mm, 7 mm and 10 mm). At the thicker end of the GTB, a kind of clamp was modelled, which can be clipped to an existing branch. This clamp improves stability on the tree and simplifies attachment by means of breathable adhesive tape.

Since the GTB has a uniform surface texture in strong sunlight, line-shaped patterns, as shown in Figure 2 (b), were painted on the GTB. The contrast is thus closer to the contrast of natural branches. The manually drawn patterns work similarly to the principle of structured light projection (Luhmann, 2023). It is expected that the pattern will improve the photogrammetric reconstruction of the GTB.

**3.2.3 Measuring Strips:** To ensure that actual shoot elongation could be validated against reference measurements, five to six measuring strips per tree were attached to every tree at different heights. Similar ruler-based methods are used for the direct measurement of shoot elongation (Schiestl-Aalto et al., 2013). The method with attached measuring strips has not yet been implemented in this way or mentioned in studies. To our knowledge, the use of attached measuring strips in this form has not yet been reported in the literature. The advantage of this measurement method is that measurements are always referenced to the same zero point, thereby minimizing reading errors compared to handheld rulers. During installation, care was taken that strips were only attached to terminal shoots with the reference mark positioned at the terminal bud. Figure 3 shows the position of the shoot apical meristem (SAM)inside the bud. The SAM is growing zone from where new leaves and internodes emerge (Baumgarten, 2022b).

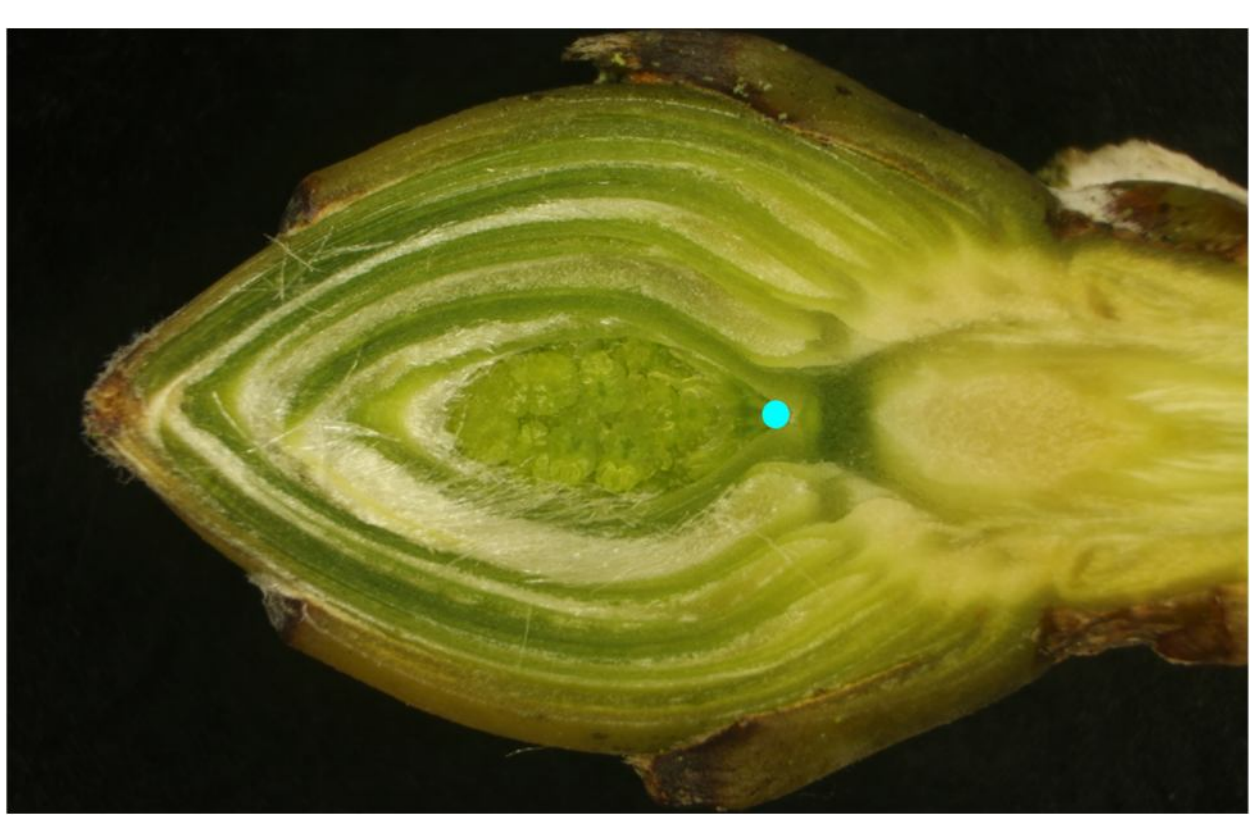

Figure 3. Microscopic Lateral section of a bud under the microscope with preformed leaves and flower structures. The base of the shoot apical meristem (SAM) is marked (Baumgarten, 2022a)

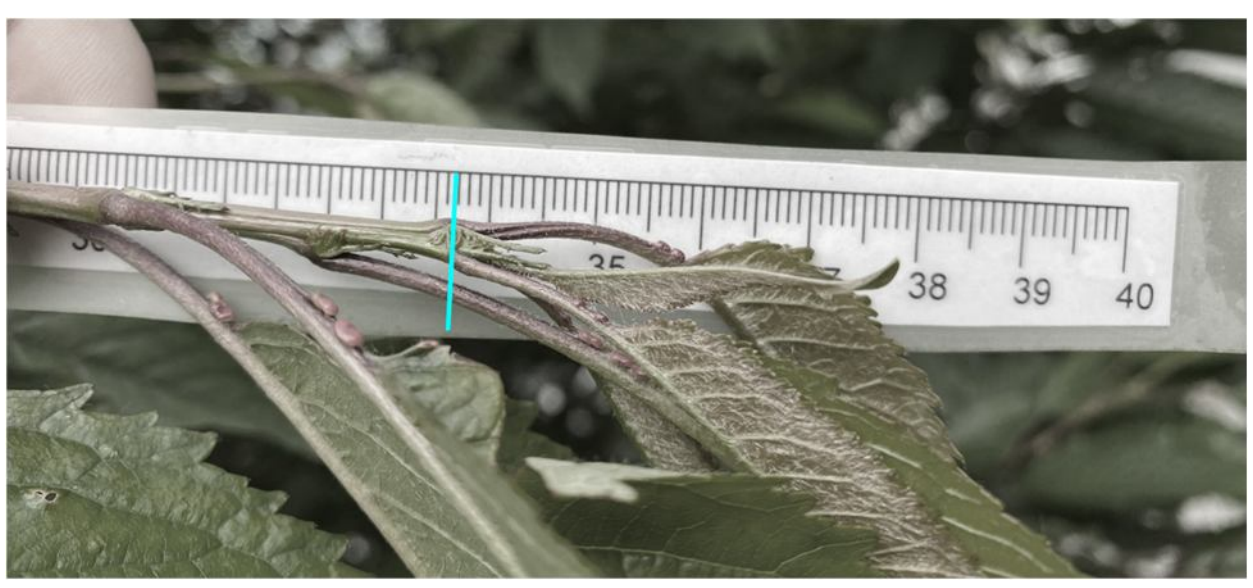

Figure 4. Shoot with the reading point marked at the origin of the innermost visible leaf

The reading of the measuring strips was carried out weekly so that a data set with high temporal resolution could be created. When recording shoot lengths, care was taken to ensure that the length is correctly read at the origin of the innermost visible leaf (see Figure 4). For the measuring strips, a reading accuracy of ±2 mm was assumed.

## 4. Methods

### 4.1 Data Acquisition

To obtain the best possible results, the sensors of different UAVs were first tested for their suitability to detect the smallest shoots. This is followed by a discussion of the primary image acquisition strategy and by alternative data capturing techniques and scenarios.

**Photogrammetric Platform and Sensor Evaluation:** The following three UAVs were available for photogrammetric recordings: “DJI Mini 3 Pro” (M3P), “DJI Mavic 3T” (M3T), “DJI Phantom 4 Pro V2” (P4P). The M3P has sensor dimensions of 9.65 x 7.2 mm, a pixel size of 1.2 µm and a resulting image size of 48 MP. The M3T features a small sensor with a size of 6.4 x 4.8 mm, a pixel size of 0.8 µm and an image size of 48 MP. The P4P’s sensor measures 13.2 x 8.8 mm, has a pixel size of 2.4 µm and a resulting image size of 20 MP. To achieve the best possible result, the three UAVs and their sensors were compared using an ISO-12233 Resolution Test Chart. After initial tests, the decision was made not to use the M3T, as the image resolution of its RGB sensor is significantly lower than that of the other two UAVs due to the smaller pixel size. Figure 5 depicts the test chart images of the M3P (a) and the P4P (b) – both obtained with a pixel size in object space of 1 mm. A visual comparison shows that the P4P has a slightly sharper image.

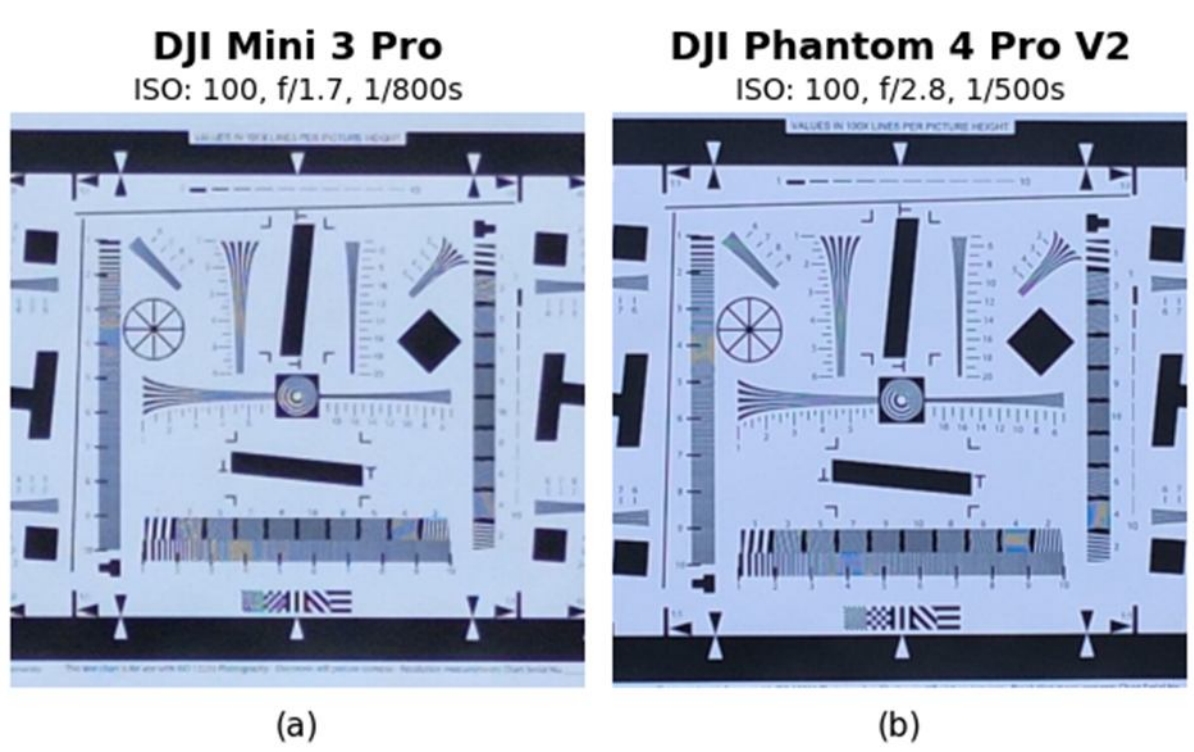

Figure 5. Comparison of test chart images acquired with the M3P with 48MP (a) and P4P with 20 MP (b).

Due to the significantly better visual results of the M3P and P4P compared to the M3T, these two UAVs were used for the recordings. In addition to the better sharpness, the P4P also offers the important possibility to fix the focus and to adjust the aperture manually. The M3P is used despite a slightly inferior sensor and blurrier images, as the weight below 250g permits flights with hardly any operational restrictions – other than the heavier P4P.

**4.1.1 Photogrammetric Image Acquisition Strategy:** Since the smallest tree shoots have a diameter of approx. 3 mm, the requirements for photogrammetric recording are correspondingly high. According to the sampling theorem of Shannon (1949), the sampling frequency must be at least twice as high as the highest frequency occurring in the signal to allow for a complete reconstruction. Regarding photogrammetry, according to Luhmann (2023), this means that the pixel size in object space, according to the Kell factor, should correspond to about one third of the smallest object size to be recorded, resulting in a pixel size of 1 mm. In addition to the object resolution, a certain depth of field in the range of 3 m to 6 m must also be achieved so that all branches in the recording area are sharply imaged. This requires fixing the focus and manually adjusting the aperture and trying to achieve the widest possible depth of field range in each case. All calculated parameters are shown in Table 1.

| Image Acquisition Parameters | DJI Mini 3 Pro (M3P) | DJI Phantom 4 Pro V2 (P4P) |
|---|---|---|
| Max. distance to tree crown | 5.6 m | 3.6 m |
| Max. pixel size (object space) | 0.5 mm | 0.8 mm |
| Number of rings | 5-9 | 11-13 |
| Number of images per ring | 59 | 42 |
| Number of images per flight | 293 - 528 | 461 - 545 |
| Separation of rings | 0.6 - 1.2 m | 0.5 - 0.5 m |
| Spacing between images | 0.3 - 0.6 m | 0.5 - 0.5 m |
| Depth of field | 0.83-3.20 m | 39.55 m - ∞ |

Table 1. Resulting parameters and values of the flight planning calculation for the M3P (portrait mode) with 48MP and the P4P (landscape mode) with 20MP.

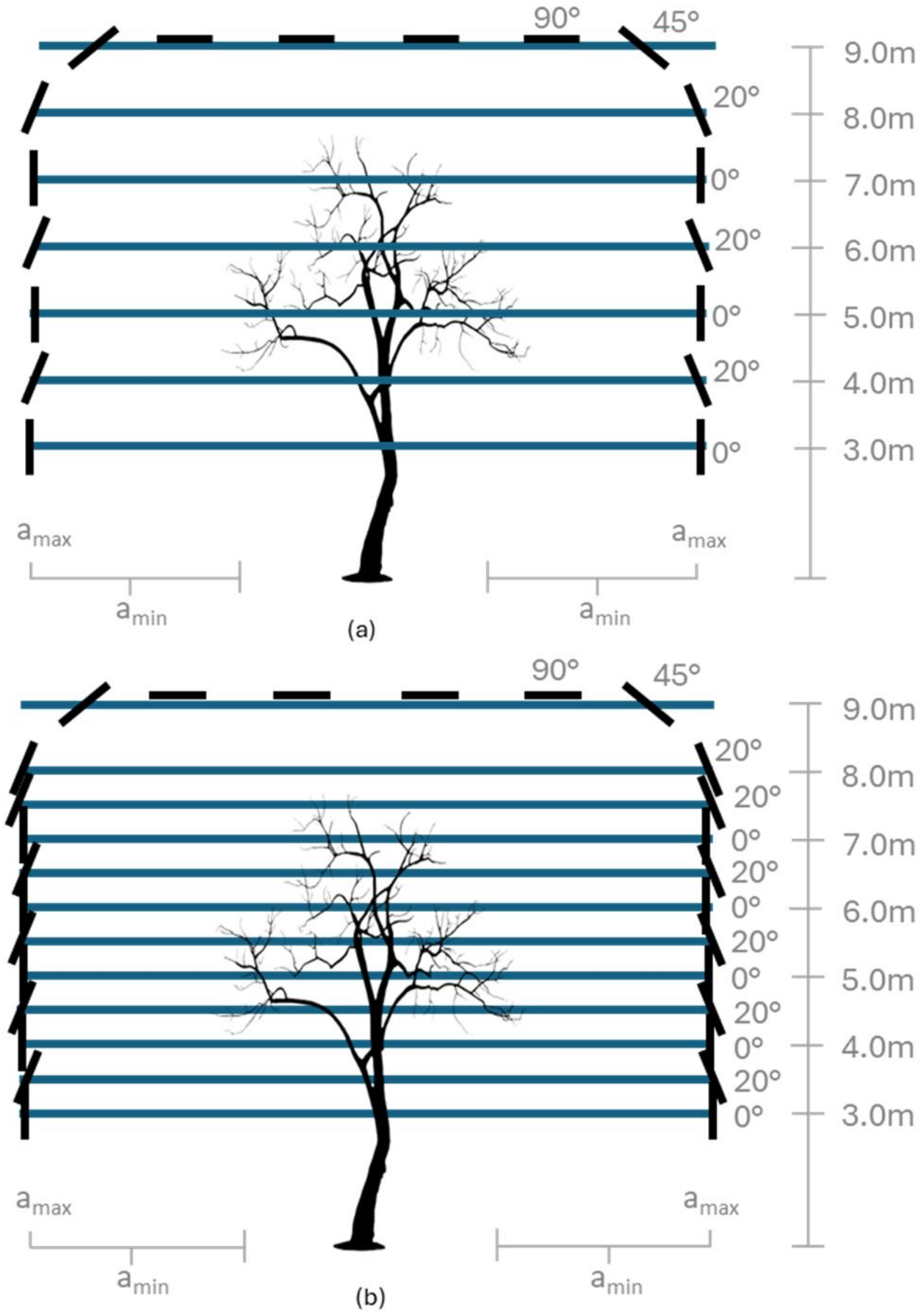

Figure 6. Visualization of the image acquisition patterns for the DJI M3P (a) and the DJI P4V2 (b). The camera tilt (gimbal pitch) is alternately set to 0° (horizontal) or 20° per circle.

To meet the required spatial resolution, flights with several rings were carried out at different heights around the tree, as in the investigations of Scher et al. (2019) and Gatziolis et al. (2015). After initial test recordings and evaluations, it was found that slightly inclined oblique images provide a better insight into the tree structure and result in more points within the treetop. Based on this finding, the rings were flown alternately at 20° and 0° and supplemented with a ring of 45° and a nadir flight above the treetop (see Figure 6). The trees in the FHNW study area have a vertically dominant shape, which suggests that images are taken in portrait format to make better use of the vertical image dimension. At the same time, the larger visible ground area in the image is expected to result in a higher number of stable tie points for image orientation. Since not all UAVs support portrait format, the flight with the P4P was planned in landscape mode, which results in a smaller distance between the individual rings. For the flights, at least 10 control points at different heights were defined per tree.

Due to neighbouring trees, limited open space and special requirements for flight planning, the individual flights had to be carried out manually. This led to differences in the number of acquired images and in a variability in object distance between individual images. In addition, the strong gusts of wind around the FHNW campus and the downdraft of the UAVs had an influence on the movements of the branches and thus on the quality of recordings and later reconstruction.

**4.1.2 UAV LiDAR:** In addition to the photogrammetric approaches, UAV-based LiDAR data acquisition was also tested. Several test flights were carried out with the Leica BLK2FLY. Data was acquired by manually flying spiral patterns around the trees. The resulting point clouds had a markedly lower point density than those of the reference data. The dispersion or noise was also very high. Branches with a diameter of 1 to 2 cm were captured with the BLK2FLY with a diameter twice as high. From this it was concluded that it is hardly possible to record the shoot lengths of branches in the millimetre range with this type of UAV LiDAR. Consequently, this measurement method was not pursued further within the scope of this project

**4.1.3 Image Acquisition with CraneCam:** The CraneCam system by Terradata AG (Terradata, AG 2024), which was developed for construction site monitoring as part of the BMETRY research project, offers a promising alternative for long-term photogrammetric multi-view data acquisition. The system consists of two (or more) cameras (see Figure 7), which are permanently mounted to the boom of a construction crane and can be aligned according to the specific project requirements. All cameras are triggered simultaneously at a certain frequency during the rotation of the crane boom and thus provide a very high overlap in direction of rotation.

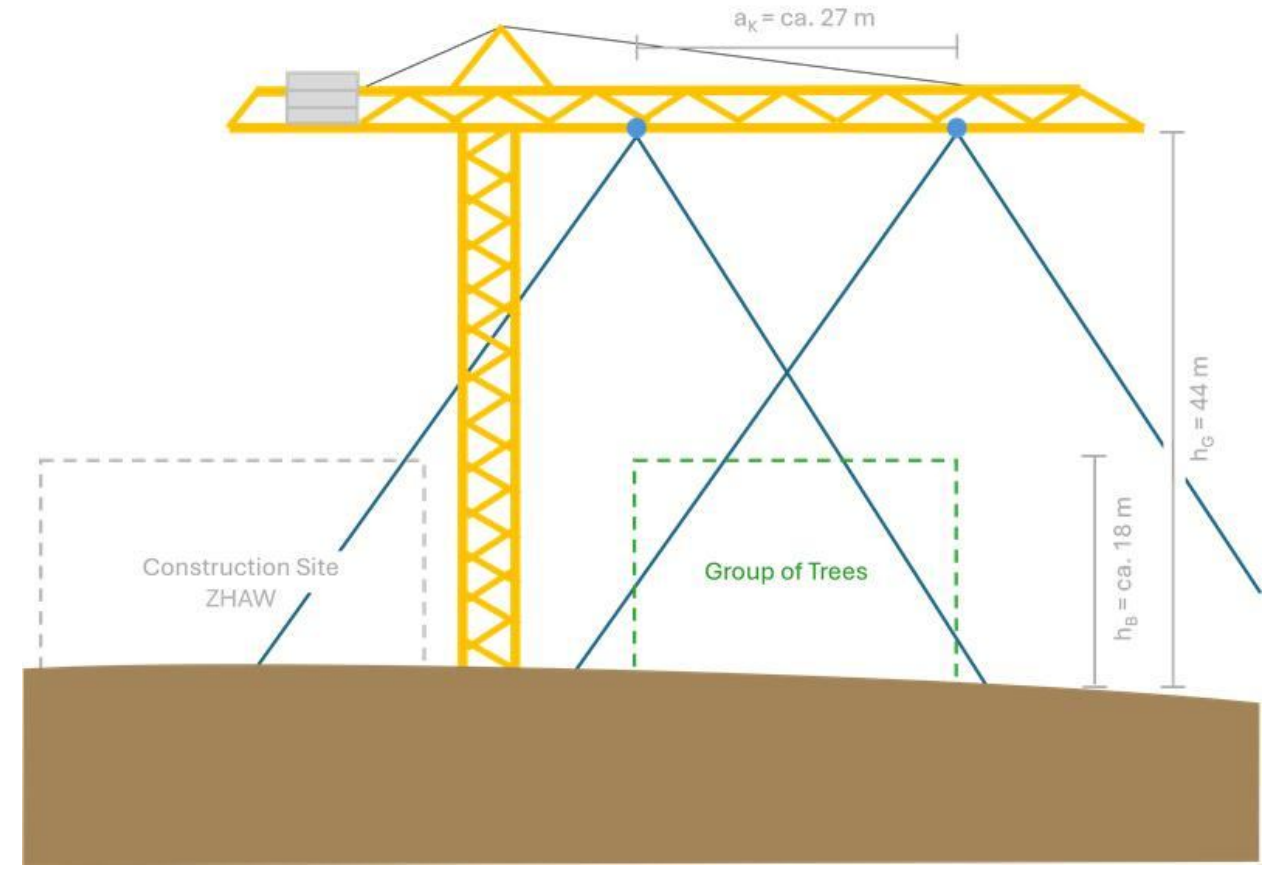

Figure 7. Schematic view of CraneCam system setup at the test site 2 in the city of Winterthur.

The CraneCam system used on test site 2 consists of two "Canon EOS M6 Mark II", each with a "Sigma Contemporary 16 mm" lens. The calculation of the object resolution shows that a resolution of 5.2 mm is obtained at the height of the treetop and 8.8 mm on the ground. Although the requirement for an object resolution of 1 mm is not met, this approach will be pursued further due to a possible use in the Hölstein Forest Laboratory. Images were taken at three different time periods, each with slightly different camera settings and recording frequencies. However, no LiDAR reference measurements could be acquired for this second study area.

**4.1.4 UAV Measurement Campaigns:** In total, 26 UAV flight campaigns were carried out on test site 1 (FHNW Campus) between 21 February and the 8 May 2025 (see Figure 8). Of these, seven missions covered a Norway maple (Acer platanoides) with the M3P and five a winter lime (Tilia cordata) also with the M3P. The main focus was placed on an Ornamental Cherry (prunus serrulate 'Shirofugen') which was observed in 11 missions using the M3P UAV and three missions using the P4P. These UAV campaigns were complemented with a total of 13 TLS reference scanning campaigns.

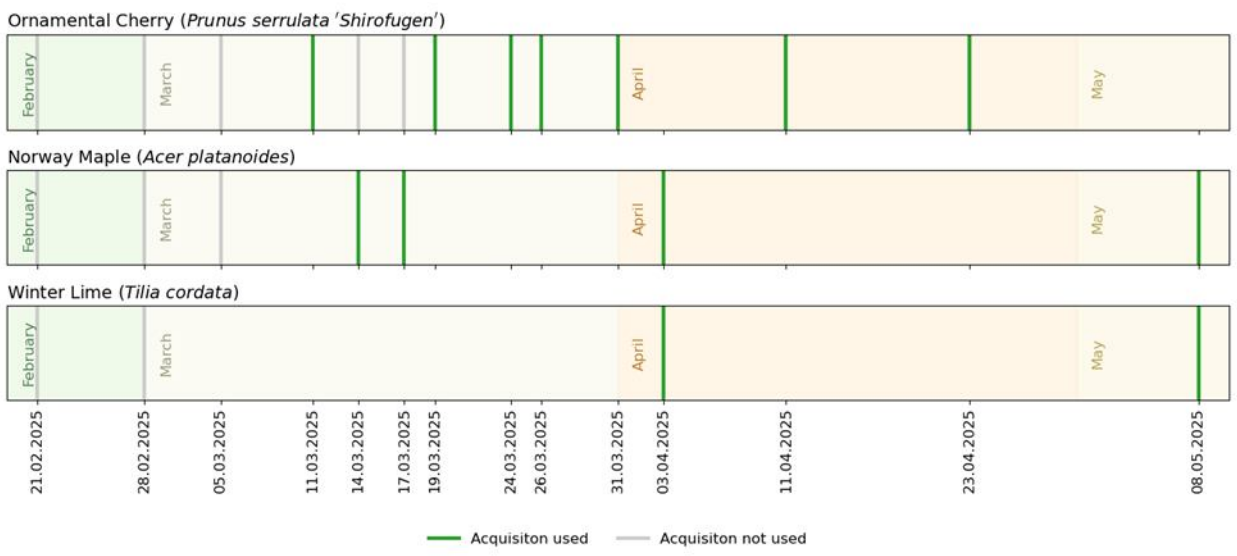

Figure 8. Overview of UAV flight campaigns on test site 1 (FHNW Campus) between 21February and 8 May 2025. Campaigns used in the evaluations are marked in green.

### 4.2 Photogrammetric Data Processing

Photogrammetric data processing involved the following main steps: evaluation of a suitable SfM-MVS software, determination of an optimal image alignment and georeferencing strategy, evaluation of dense cloud generation and results, and first skeletonization experiments.

**4.2.1 Software Evaluation:** In an initial step, three SfM-MVS software solutions were evaluated for their suitability for the specific task of these investigations: Pix4D Matic, ESRI Drone2Map and Agisoft Metashape. Pix4D was excluded at an early stage and not further considered due to systematic effects in the image alignment results and in systematic, primarily vertical offsets within the subsequent sparse point clouds. The remaining two software packages showed a similar overall performance. Drone2Map ranked better in terms of reprojection errors and RMSE for GCPs and check points. A direct comparison for the M3P (DJI Mini 3 Pro) flight mission of the 26.03., for example, yields an average reprojection error of 1.2 pixels with Drone2Map and 1.7 pixels with Metashape. These values are significantly higher than those expected from high-grade image sensors, which are not available with light-weight low-cost UAVs such as the M3P.

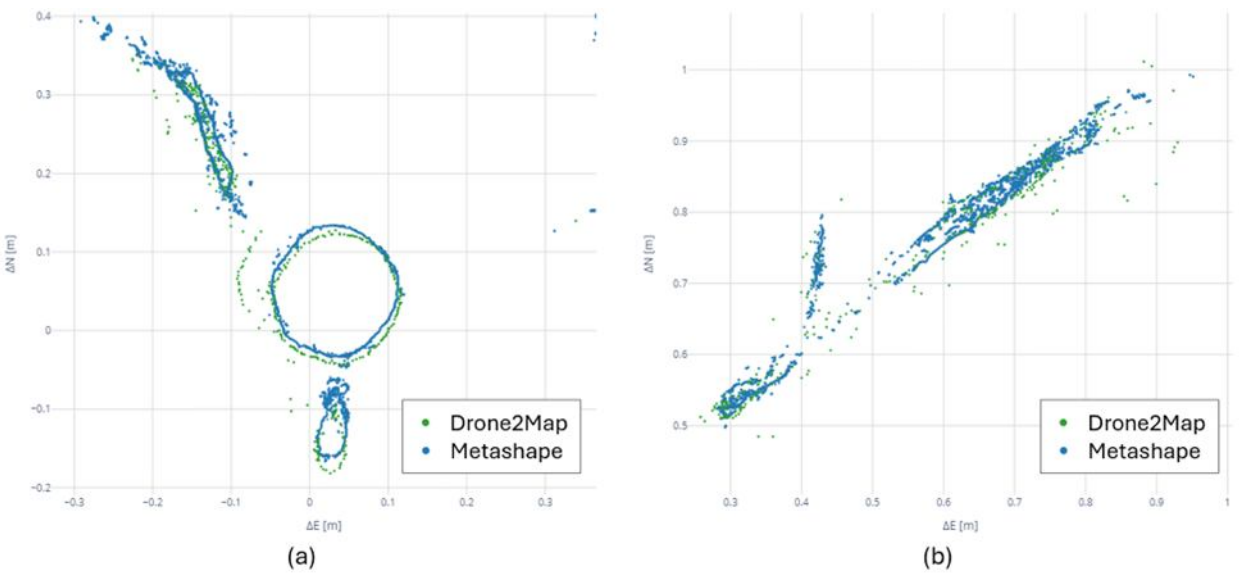

Figure 9. Software evaluation: comparison of point clouds from ESRI Drone2Map and Agisoft Metashape. Horizontal slices with thickness of 2 mm. Flight missions with M3P on 11.03.25 (a) (trunk & branch) and on 26.03.25 (b) (branch).

With respect to generated dense point clouds, there is a vast difference between the two evaluated software solutions. MetaShape delivers a 13-fold higher number of points (approx. 45.5 Mio. points vs. 3.5 Mio. points per tree). The visual comparison of horizontal sections within the point cloud shown in Figure 9 also illustrates that MetaShape reconstructs more detailed branch structures with lower noise and higher point density than Drone2Map. MetaShape also offers significantly more setting options, which is why MetaShape was selected for the subsequent evaluations.

**4.2.2 Image Alignment and Georeferencing:** Monitoring the growth of individual tree-shoots over an extended period of time, as well as various comparisons with reference data, place very high demands on the accuracy of image alignment and georeferencing. To ensure optimal image alignment – and 3D reconstruction results – with the given low-cost sensors and image acquisition patterns, a thorough evaluation of image alignment parameters in Metashape was conducted. The resulting parameters, which were subsequently used for the evaluation of the entire time series, are listed in Table 2.

| **Image alignment setttings** | |
|---|---|
| Accuracy | Highest |
| Generic Preselection | Yes |
| Reference Preselection | Estimated |
| Key point Limit | 60000 |
| Guided image matching | No |

Table 2. Summary of optimal image alignment parameters in Agisoft Metashape

While the above parameters yield optimal image alignment results, GCPs play a fundamental role a) in co-registering multiple measurements epochs, b) in stabilizing image alignment and c) in allowing highly accurate comparisons with the TLS reference scans. Throughout the measurement campaigns a combination of coded targets and of selected natural GCPs were used. The latter ensure the long-term availability of control point data over the project duration of several months. GCPs were measured with a total station and tied to the geodetic control network of FHNW. The 3D accuracy of all GCPs is better than 10 mm.

**4.2.3 Dense Point Cloud Generation and Evaluation:** The SfM-based image alignment and sparse point cloud creation were followed by the dense point cloud generation. Table 3 lists the processing parameters for dense cloud generation in Metashape, which resulted from the respective evaluation and were subsequently used for processing the entire time series.

| **Build point cloud settings** | |
|---|---|
| Source data | Depth maps |
| Quality | Ultra High |
| Depth filtering | Mild |

Table 3. Summary of key parameters for dense point cloud generation in Metashape

In a post-processing step, the resulting point clouds were then each trimmed with a Python script and cleaned up in CloudCompare to isolate the investigated tree. Examples of resulting dense point clouds are shown in Figure 1 with epochs 24.03. (left), 24.03. (middle) and 23.04.25 (right).

Subsequently, the dense point clouds were examined for accuracy, noise and completeness using the following types of reference data: the 3D-printed ground-truth branch (GTB) which had been fitted to the tree (see Figure 2) and TLS scans for the respective epochs.

Different quality aspects and metrics of the photogrammetric dense point clouds were evaluated using the following methods:

a) Reconstruction accuracy & resolution: comparison with the reference dimensions of the GTB
b) Noise: comparison with TLS reference scans using cloud-to-cloud metrics (C2C)
c) Completeness: 1) evaluation using C2C with an inverted order (distances of photogrammetric points to TLS reference point cloud) together with a tolerance limit and 2) reprojection of the dense point cloud into image space for a visual inspection.

### 4.3 Skeletonization Approach

Skeletonization of the point clouds is a promising approach to determine the absolute growth of shoot lengths. As part of this project, first investigations were carried out to evaluate the suitability of the UAV- and CraneCam-based photogrammetric point clouds for complete skeletonizations of entire trees. Following a short evaluation of existing skeletonization approaches for trees, TreeQSM by Raumonen & Åkerblom (2025) was selected for first experiments. TreeSQM works by segmenting the point cloud and then creating a cylinder model, from which the cylinder axis is extracted as a skeleton. The comparison of skeletons for two epochs is expected to serve as a basis for determining absolute shoot elongation.

## 5. Experiments and Results

The main result of each UAV campaign was a georeferenced and textured dense point cloud of an entire tree at the respective epoch as illustrated in Figure 1. In the following sections, the experiments investigating different aspects of the image-based reconstruction and their respective results are presented.

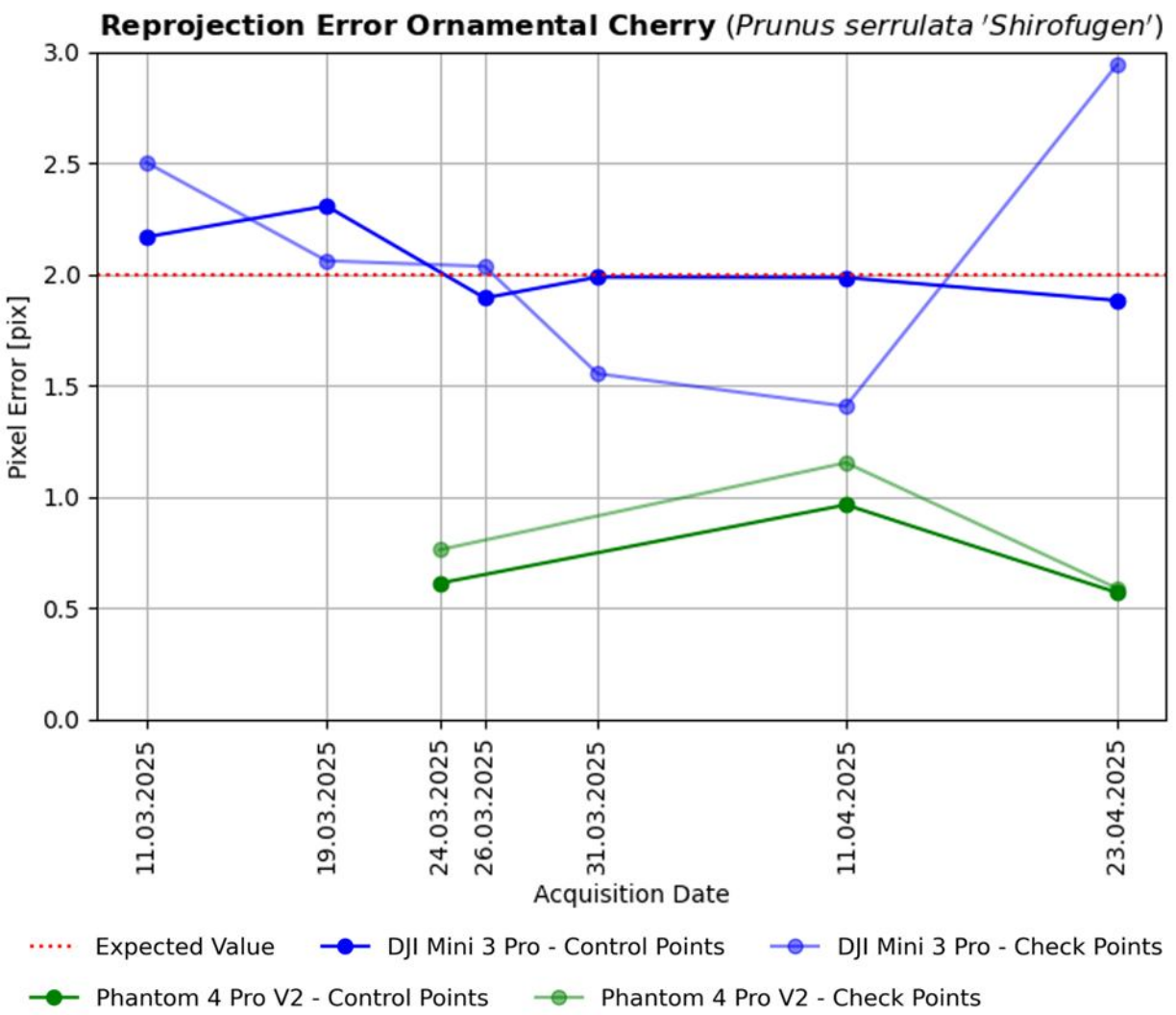

Figure 10. Visualization of the reprojection errors for different flight campaigns and both UAVs (M3P in blue; P4P in green)

### 5.1 Image Alignment and Georeferencing

Optimizing the image alignment parameters is a crucial aspect in getting results with high quality. But the even bigger impact can be made with a bigger camera sensor and adaptable and fixable camera parameters. In contrast to the extremely light-weight and low-cost M3P sensor, the P4P allows to fix the focal distance and to set the aperture priority. This ensures more stable camera parameters and leads to significantly better results in image alignment as can be seen in Figure 10. The average reprojection error improves from 2.0 pixels for the M3P to 0.8 pixels for the P4P. 3D point accuracy evaluated based on check points also shows an improvement from 6.5 mm to 5.1 mm respectively.

### 5.2 Evaluation of Point Cloud Quality

To assess the photogrammetric results, the point clouds were checked against the reference data. They were examined for noise, completeness and differences with respect to the GTA.

**5.2.1 Point cloud noise:** Cloud-to-Cloud (C2C) algorithms in CloudCompare were used to evaluate noise characteristics and subsequent completeness. To filter gross errors, the maximum distance of the C2C comparison was set to 10 cm. The noise comparison between the P4P and the M3P already shows major differences. While the P4P point cloud has hardly any noise (see Figure 11 (a)), the M3P shows strong noise in the upper area of the treetop (Figure 11 (c)). Even though the two data sets are not from the exact same day, this should not have a significant impact on the findings, as growth was barely taking place at that time. Individual red areas exceeding the maximum distance of 10 cm can also be seen in the P4P point cloud (Figure 11 (a)). These areas show the reference measuring strips, which had been attached to the tree between the reference scans and the UAV data acquisition.

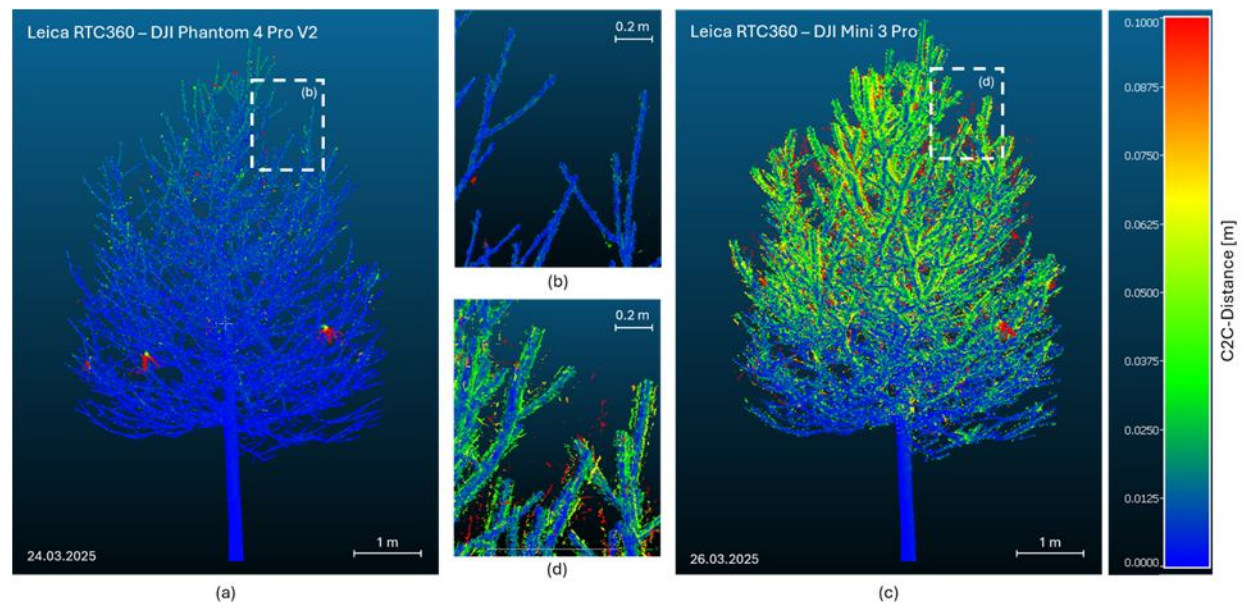

Figure 11. Cloud-to-cloud comparison between the TLS reference point cloud of March 24, 2025 and the point clouds of the P4P on March 24 (a) and of the M3P on March 26 (c) including respective detail views (b) and (d).

The histograms in Figure 12 illustrate the different distribution characteristics of the C2C differences for the two UAVs with a mean value of 4.5 mm and a SD of 6.7 mm for the P4P (a) and a mean of 11.6 mm and SD of 16.6 mm for the M3P (b).

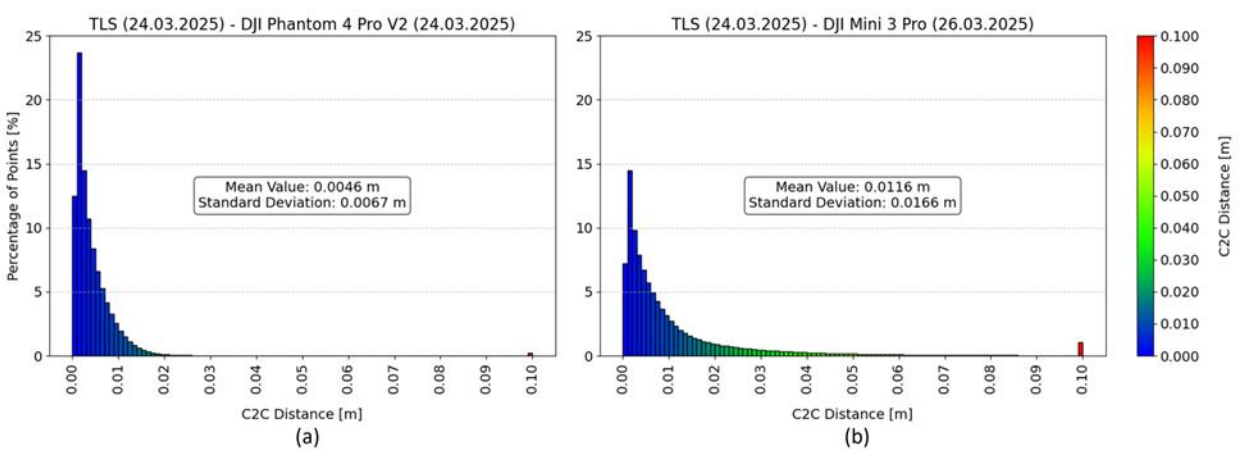

Figure 12. Histograms of the C2C distances including mean and standard deviation. Comparison of TLS with P4P (a) and with M3P (b)

**5.2.2 3D Reconstruction Accuracy and Resolution:** The ground-truth branch (GTB) allows to evaluate both the accuracy and the resolution of photogrammetric reconstructions under real-world conditions. The comparison of the various photogrammetric point clouds from both UAVs with the GTA was performed based on sample measurements. These measurements were carried out in the original point clouds and in the ones filtered by confidence values. Figure 13 shows filtered point cloud segments reprojected into the original images for the P4P UAV (a) and for the M3P (b).

The box plots in Figure 14 show the differences between the diameters derived from the point clouds and the reference diameters of the GTA. The diameters observed in the point clouds are generally overestimated, in case of the P4P by an average of 3 mm and in case of the M3P by 7 mm. The different gradations in diameter have no significant influence on the obtained differences. The length is completely reconstructed in all data sets, which leads to the conclusion that the smallest branches can be reconstructed with the defined pixel size in object space of 1 mm.

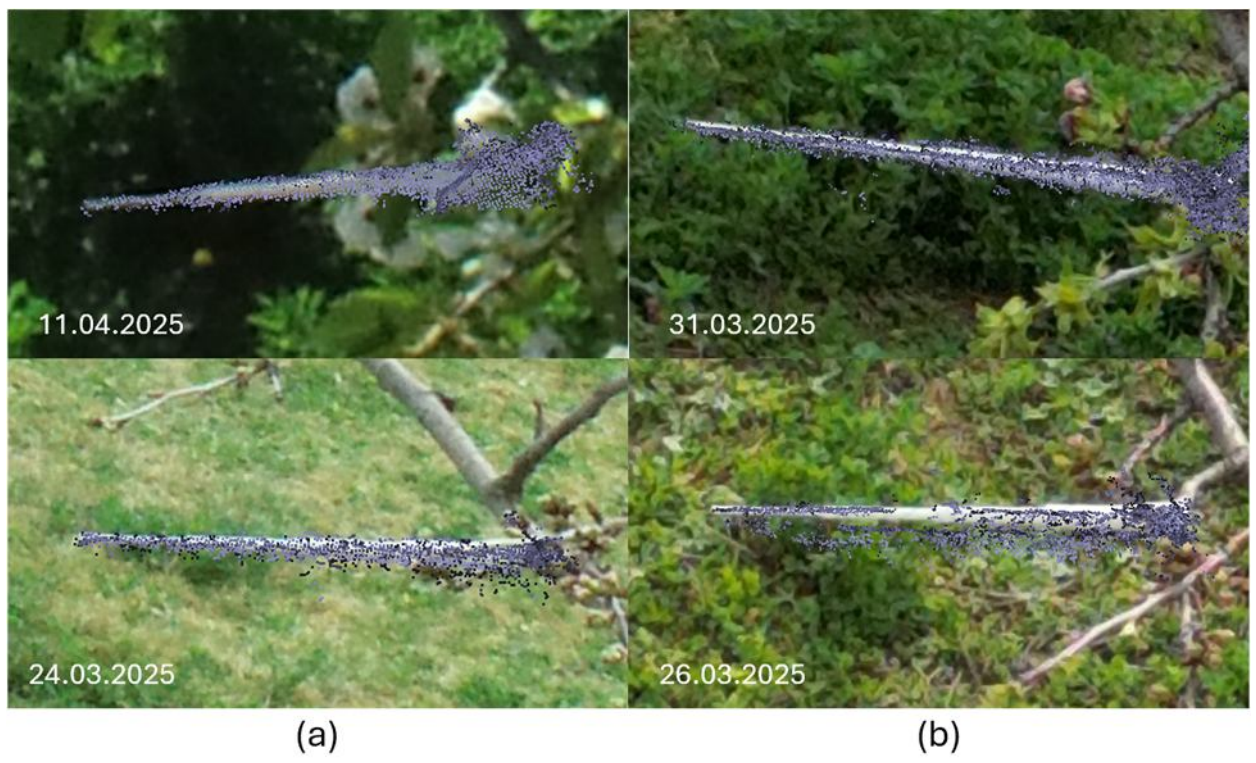

Figure 13. Visual quality control of point cloud scattering based on the GTB including reproduced point cloud. Excerpts of P4P point clouds (a) and M3P point clouds (b) from different epochs.

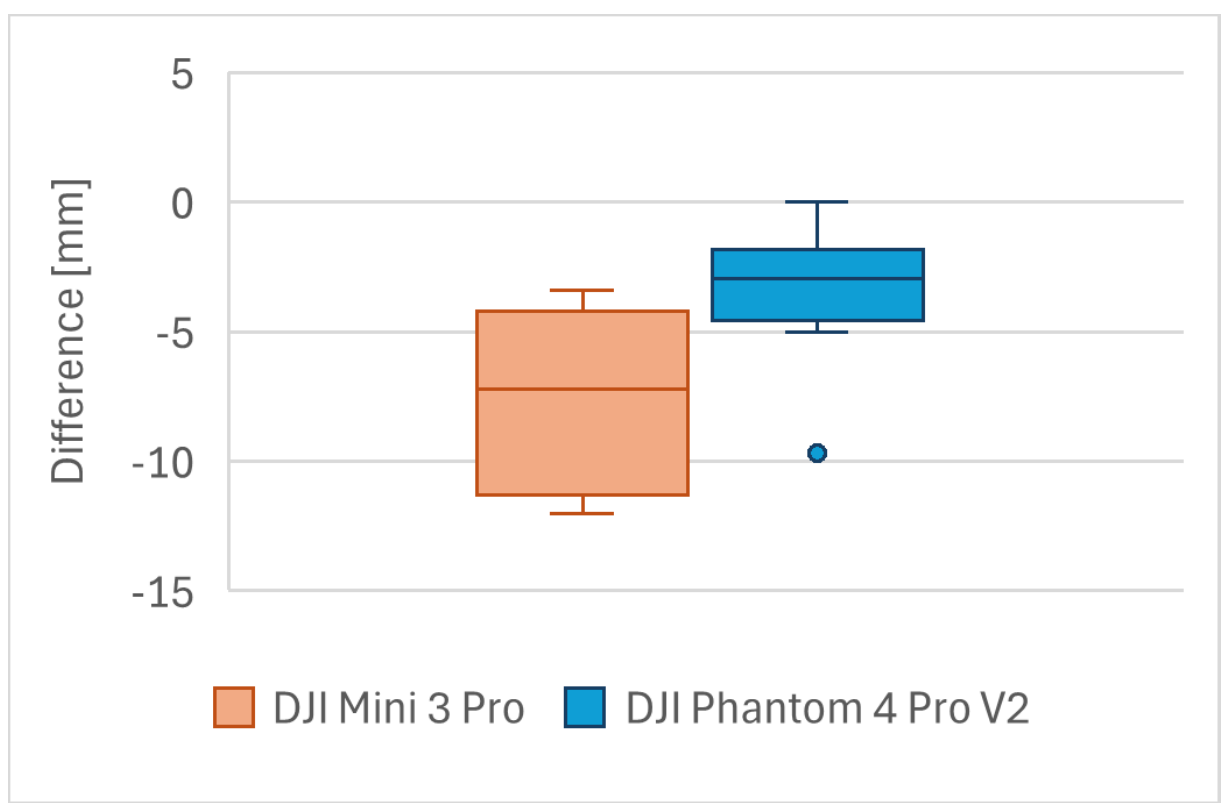

Figure 14. Differences of GTB branch diameters derived from the dense point clouds (M3P in orange; P4P in blue) and the reference values of the ground-truth branch (GTB).

**5.2.3 Point Cloud Completeness:** A C2C comparison with the TSL reference scans was also carried out to evaluate the completeness of the photogrammetric point clouds. The results of the C2C comparison can be seen in Figure 15. Generally, the comparison shows only very small differences between the data of the P4P and the M3P. However, several minor deviations are visible over the entire crown. This is due to the noise in the TLS reference point cloud. Since the point cloud of the M3P has stronger noise than that of the P4P, shorter C2C distances result compared to the TLS point cloud. By filtering all points with a C2C distance of more than 5 cm, incomplete shoots can be identified. In the point clouds of both UAVs, individual incomplete shoots are visible and are considered in the subsequent sample count.

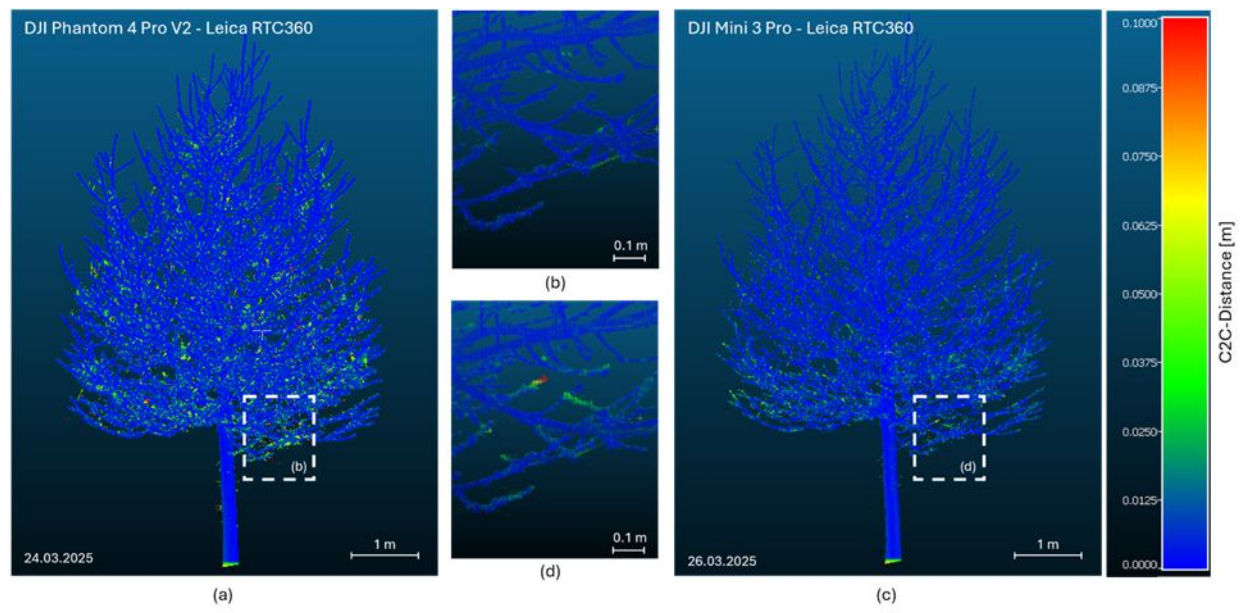

Figure 15. Cloud-to-Cloud differences between the TLS reference point cloud (24.03.25) and the P4P point cloud (24.03.25) (a) and the M3P point cloud (26.03.25) (c). Detail views (b) and (d) illustrating incomplete shoots.

To obtain an effective percentage value for completeness, samples were again collected manually. To do this, the point clouds were projected into the images in MetaShape (see Figure 16) and individual branches were visually assessed to see whether they had been completely reconstructed. This resulted in a slightly higher completeness of 98% for the P4P (based on 234 samples) over 92% for the M3P (based on 214 samples).

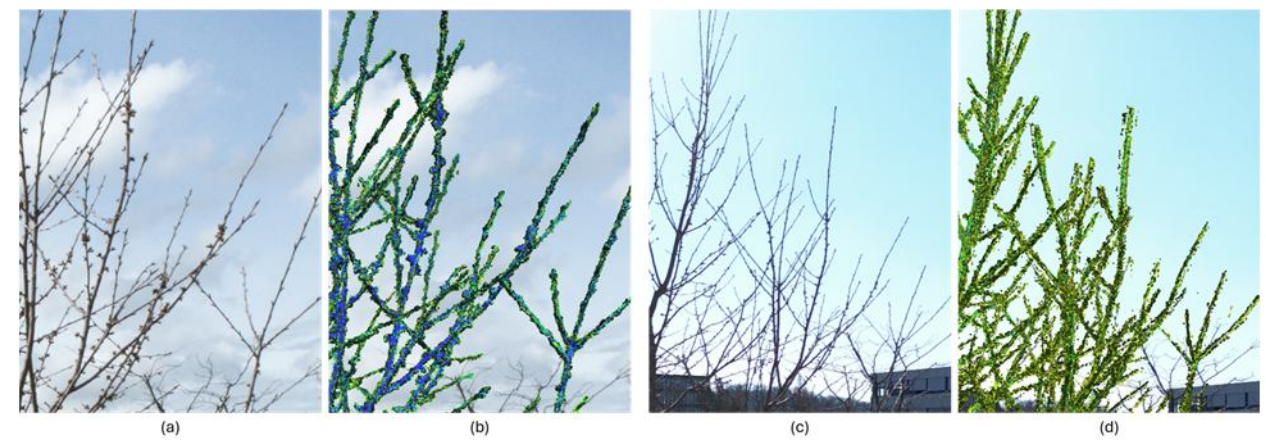

Figure 16. Reprojection of the photogrammetric dense point cloud into the original UAV images illustrating the highly detailed reconstruction and the very high degree of completeness. Results from P4P: (a) and (b) and from M3P: (c) and (d).

The experiments in test area 2 with the CraneCam imagery also showed a detailed 3D reconstruction of branch structures and shoots as shown in Figure 17, despite the significantly lower ground sampling distance than test site 1. In the absence of independent reference data, the 3D reconstructions in test area 2 were primarily assessed visually.

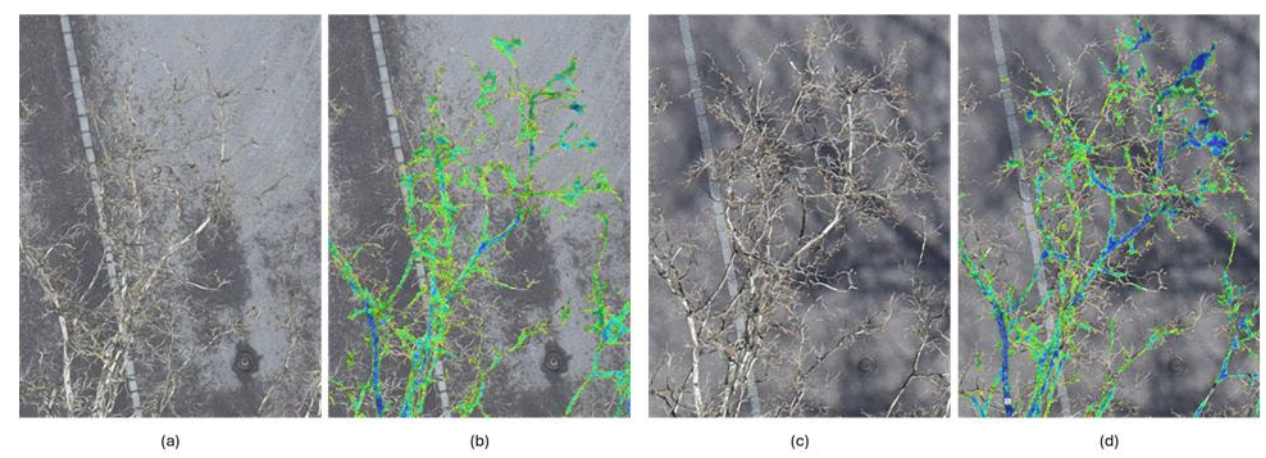

Figure 17. Reprojection of the photogrammetric dense point cloud into the original CraneCam images for test site 2.

## 5.3 First Skeletonization Results

Within the scope of this work, first experiments were carried out to investigate a) the suitability of the photogrammetric point clouds for a skeletonization of entire trees and b) the feasibility of a complete skeletonized description as basis for future absolute shoot growth analysis. Figure 18 shows a tree skeleton (a) obtained using TreeQSM by Raumonen & Åkerblom (2025) and a superimposition of the skeleton with the original point cloud (b). While the first results show a reasonably good overall representation of the tree, the skeleton still contains a fairly large number of blunders, for example, disconnected edge sections or incorrect edges not representing a physical branch or shoot. In the current state, the skeletons are not yet suitable for the envisaged complete analysis of absolute shoot growth.

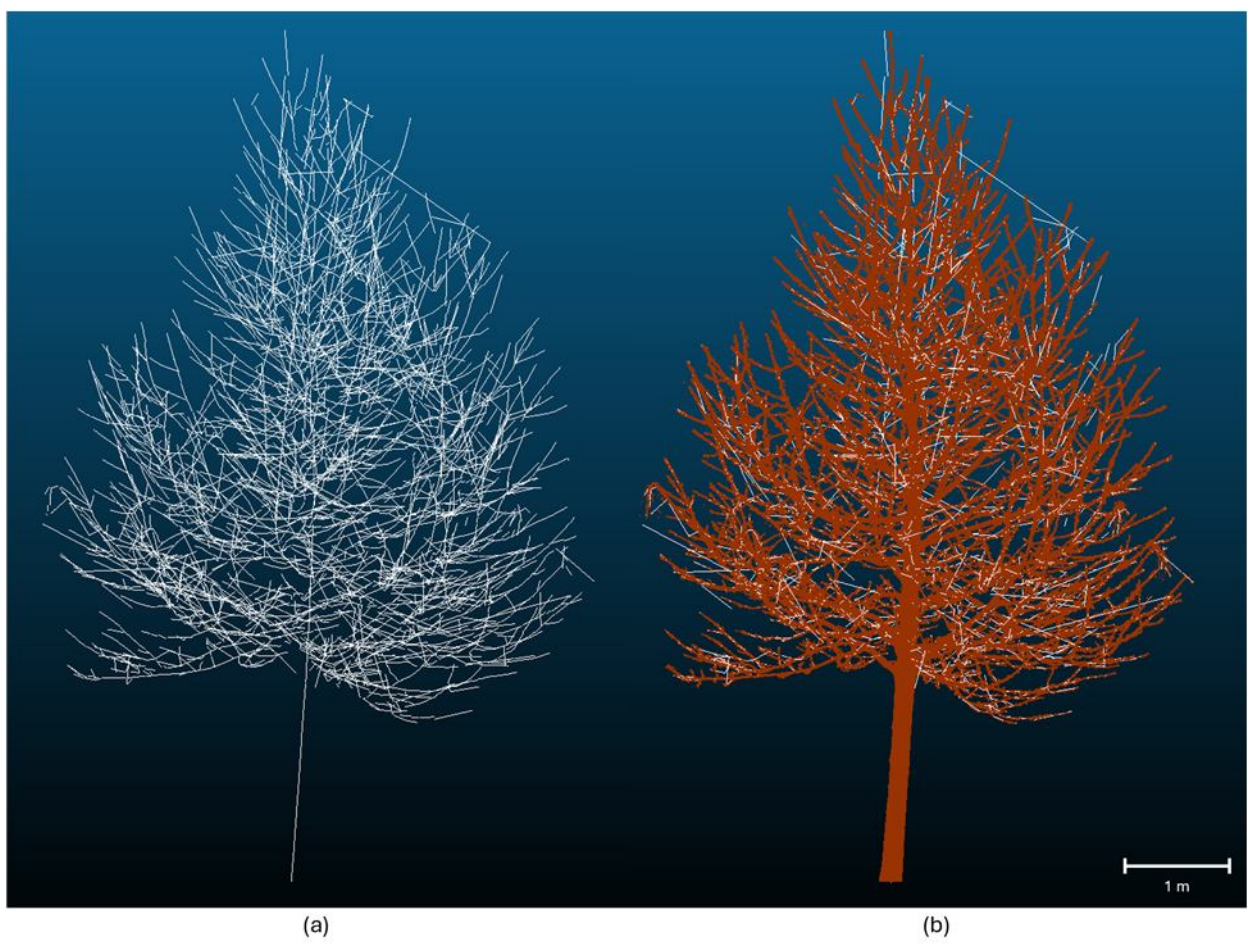

Figure 18. Extracted tree skeleton (a) and skeleton superimposed with the input point cloud (b) (P4P, 24.03.25)

## 6. Discussion

The goal of the presented investigations and results was to determine whether accurate and complete 3D reconstructions of entire deciduous trees using low-cost UAV photogrammetry is feasible. The following aspects influence the results:

- UAV sensors and 3D reconstruction accuracies: The comparison between different UAV platforms and sensors showed that accurate and complete reconstructions are possible, even with consumer-grade UAVs such as the DJI Mini Pro 3 weighing less than 250g. However, the investigations also show a 25% improvement in 3D point accuracy to approx. 5 mm when using a semi-professional UAV platform such as the DJI Phantom 4 Pro V2 with a camera offering better geometric resolution and control over focus and aperture.
- Completeness: With both types of UAVs individual trees could be reconstructed in the form of dense point clouds with a very high level of completion. The achieved completeness was 98% using the P4P UAV and still 92% using the M3P.
- Ground-truth branch: The introduction of a novel 3D-printed ground-truth branch (GTB) allowed to evaluate both the geometric accuracy and resolution of the 3D reconstruction. With the target GSD of 1 mm, both systems were capable of reliably detecting the thinnest shoots with a diameter of just 3 mm. However, the P4P outperformed the M3P in the estimation of the shoot thickness.
- Operational challenges: Acquiring an entire time series of photogrammetric flight missions proved very ambitious for several reasons. Firstly, narrow spacing and close-up sensing required a manual data acquisition since there is currently no mission control software capable of flying such demanding photogrammetric flight mission. Secondly, the test area is frequently affected by strong winds, reducing the suitable time slots for successful flights and reconstructions.
- Skeletonization: The photogrammetric point clouds with their high accuracies, densities and level of completeness provide a promising base for new skeletonization approaches. However, there are several challenges such as overlapping branches and disturbing objects like growing leaves which available algorithms have not yet met.

## 7. Conclusion and Outlook

The aim of this work was to develop a method for recording and measuring the primary growth of deciduous trees under real-world conditions. The investigations showed that highly detailed 3D reconstructions of entire deciduous trees as basis for subsequent shoot growth monitoring using low-cost UAV photogrammetry is possible. Crane-based multi-camera setups for research sites such as the Swiss Canopy Crane II (Arend & Kahmen, 2019) with tailor-made camera setups are also very promising, since they are less susceptible to wind-induced branch and leaf motion.

The developed measurement and evaluation method will serve as basis for further research on the monitoring of shoot elongation at a high temporal resolution. Furthermore, the acquired time series of UAV imagery and 3D reconstructions over an entire growth season provide an ideal basis for further investigations, for example, using the latest AI-based tree skeletonization approaches, such as published by Marks et al. (2025).